\documentclass[sigconf,nonacm]{acmart}

\renewcommand\footnotetextcopyrightpermission[1]{} % suppress footnote

\AtBeginDocument{%
  }

\acmConference[ ]{ }{ }{ }
\acmBooktitle{}
\acmISBN{}
\usepackage{multirow}
\usepackage{makecell}

\usepackage{subcaption}

\begin{document}

\title{Omni-Embed-Nemotron: A Unified Multimodal Retrieval Model for Text, Image, Audio, and Video}

%%
%% The "author" command and its associated commands
\author{Mengyao Xu}
\email{mengyaox@nvidia.com}
\affiliation{%
  \institution{NVIDIA}
  \city{Santa Clara}
  \country{USA}
}

\author{Wenfei Zhou}
\email{wenfeiz@nvidia.com}
\affiliation{%
  \institution{NVIDIA}
  \city{Los Angeles}
  \country{USA}
}

\author{Yauhen Babakhin}
\email{ybabakhin@nvidia.com}
\affiliation{%
  \institution{NVIDIA}
  \city{Prague}
  \country{Czechia}
}

\author{Gabriel Moreira}
\email{gmoreira@nvidia.com}
\affiliation{%
  \institution{NVIDIA}
  \city{São Paulo}
  \country{Brazil}
}

\author{Ronay Ak}
\email{ronaya@nvidia.com}
\affiliation{%
  \institution{NVIDIA}
  \city{Sarasota}
  \country{USA}
}

\author{Radek Osmulski}
\email{rosmulski@nvidia.com}
\affiliation{%
  \institution{NVIDIA}
  \city{Brisbane}
  \country{Australia}
}

\author{Bo Liu}
\email{boli@nvidia.com}
\affiliation{%
  \institution{NVIDIA}
  \city{New York}
  \country{USA}
}

\author{Even Oldridge}
\email{eoldridge@nvidia.com}
\affiliation{%
  \institution{NVIDIA}
  \city{Vancouver}
  \country{Canada}
}

\author{Benedikt Schifferer}
\email{bschifferer@nvidia.com}
\affiliation{%
  \institution{NVIDIA}
  \city{Berlin}
  \country{Germany}
}


\begin{abstract}
We present Omni-Embed-Nemotron, a unified multimodal retrieval embedding model developed to handle the increasing complexity of real-world information needs. While Retrieval-Augmented Generation (RAG) has significantly advanced language models by incorporating external knowledge, existing text-based retrievers rely on clean, structured input and struggle with the visually and semantically rich content found in real-world documents such as PDFs, slides, or videos. Recent work such as ColPali has shown that preserving document layout using image-based representations can improve retrieval quality. Building on this, and inspired by the capabilities of recent multimodal models such as Qwen2.5-Omni, we extend retrieval beyond text and images to also support audio and video modalities. Omni-Embed-Nemotron enables both cross-modal (e.g., text → video) and joint-modal (e.g., text → video+audio) retrieval using a single model. We describe the architecture, training setup, and evaluation results of Omni-Embed-Nemotron, and demonstrate its effectiveness in text, image, and video retrieval.
\end{abstract}

%% CCS concepts - Replace with appropriate concepts for your paper
\begin{CCSXML}
<ccs2012>
 <concept>
  <concept_id>10002951.10003260.10003277</concept_id>
  <concept_desc>Information systems~Information retrieval</concept_desc>
  <concept_significance>500</concept_significance>
 </concept>
 <concept>
  <concept_id>10010147.10010178.10010179</concept_id>
  <concept_desc>Computing methodologies~Neural networks</concept_desc>
  <concept_significance>300</concept_significance>
 </concept>
</ccs2012>
\end{CCSXML}

\ccsdesc[500]{Information systems~Information retrieval}
\ccsdesc[300]{Computing methodologies~Neural networks}

\keywords{multimodal retrieval, embedding models, cross-modal search, neural information retrieval, video retrieval}

%\begin{figure}[t!]
%  \includegraphics[width=\textwidth]{omni_5.png}
%  
%  \label{fig:omni}
%\end{figure}


\maketitle

\section{Introduction} 

\begin{figure*}
  \includegraphics[width=\textwidth]{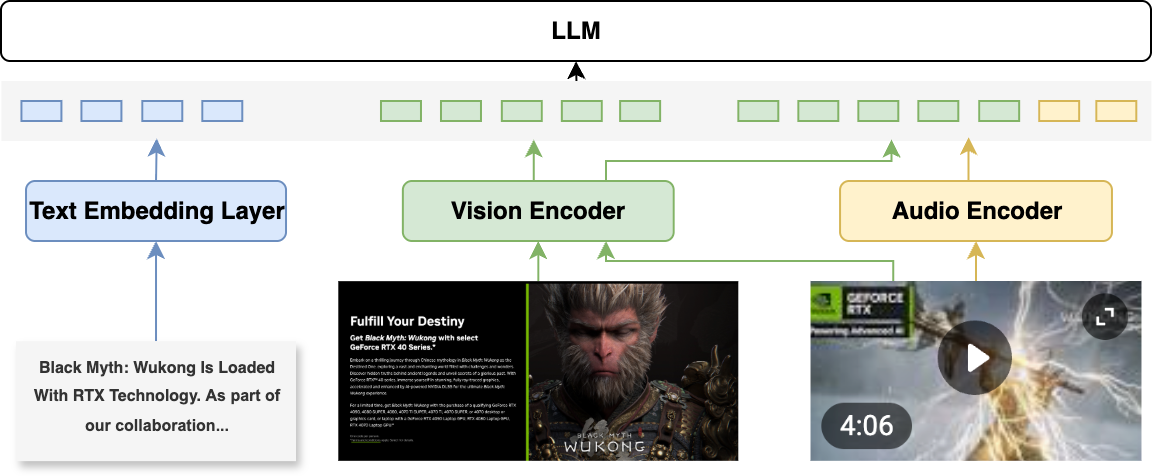}
  \caption{Multimodal Retrieval Architecture with three input modalities: text, image, and video.}
  \label{fig:omni}
\end{figure*}

Retrieval-Augmented Generation (RAG) enhances language models by integrating external knowledge retrieval, enabling models to reason over large corpora with improved factual accuracy and context awareness. Recent advancements in text retrieval have led to high-performing models such as NV-Embed~\cite{nv-embed}, NV-Retriever~\cite{nv-retriever}, Qwen3-Embedding~\cite{qwen3-embed}, and e5-mistral~\cite{e5-mistral}, which achieve strong results on benchmarks like the Massive Text Embedding Benchmark (MTEB)~\cite{mteb,mtebnew}, a widely used evaluation benchmark designed to assess the performance of text embedding models across a broad range of retrieval, classification, and clustering tasks. However these models require clean, well-formatted text input, real-world documents (e.g., PDFs, slides) often contain complex visual layouts. To address this, ColPali~\cite{colpali} proposes converting documents into images, preserving both text and visuals for retrieval. Vision-Language Models (VLMs) such as Qwen-VL~\cite{qwenvl}, LLaMA-3.1-Nemotron-Nano-VL~\cite{nemotron}, and Eagle2~\cite{eagle,eagle25} learn joint multimodal representations using encoders like CLIP~\cite{clip}, SigLIP~\cite{siglip}, and C-RADIO~\cite{cradio}. These developments motivate multimodal retrieval systems capable of understanding documents in their native visual form and learn joint text-image representations, thus there are many visual retrieval models, such as Colpali~\cite{colpali}, Llama Nemoretriever Colembed~\cite{nemoretriever-colembed}, 
Colnomic Embed Multimodal~\cite{nomic}.

Real-world content goes beyond text and images, encompassing rich modalities such as audio and video, as shown in Figure~\ref{fig:omni}. As a result, traditional unimodal retrieval methods are increasingly insufficient for handling diverse query formats and heterogeneous data sources. To address this limitation, we propose Omni-Embed-Nemotron, a unified retrieval model capable of retrieving relevant documents and media across modalities—including text, images, audio, and video. Our approach supports both cross-modal and joint-modal retrieval, enabling queries such as video-only or combinations like video+audio+images.

\section{Model}

\begin{figure}[t!]
  \centering
  \includegraphics[width=\linewidth]{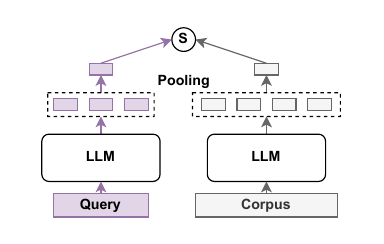}
  \caption{Bi-encoder retrieval system where both query and corpus are encoded by the same LLM, followed by a pooling layer. The resulting representations are then compared to compute a similarity score.}
  \label{fig:biencoder}
\end{figure}

Our retrieval system is built as a bi-encoder model, where queries and corpus items are independently encoded into a shared embedding space, as shown in Figure \ref{fig:biencoder}. At inference time, relevance is computed efficiently using similarity metrics such as dot product or cosine similarity. This architecture offers high scalability and supports efficient retrieval from large corpora.

A key feature of our model is its support for cross-modal and joint-modal retrieval. Queries and corpus entries can be drawn from any single modality (e.g., text, image, audio, or video) or any combination thereof (e.g., text+audio → image+video). This enables rich and flexible retrieval use cases, such as querying a video database with spoken descriptions and textual keywords, or retrieving relevant audio-visual content given a multimodal prompt.

We build our model on Qwen-Omni / Qwen2.5-Omni-3B~\cite{Qwen2.5-Omni}, a multimodal foundation model designed to process inputs across vision, language, audio, and video. 
Specifically, we utilize the Thinker backbone of the Qwen-Omni architecture, which is responsible for cross-modal understanding. We discard the Talker component, as our model focuses on retrieval tasks and does not generate audio outputs. This design choice simplifies the architecture while retaining strong multimodal representation capabilities.

\begin{figure}[t]
  \centering
  % Two side-by-side images using minipage
  \begin{minipage}[t]{0.45\linewidth}
    \centering
    \includegraphics[width=\linewidth]{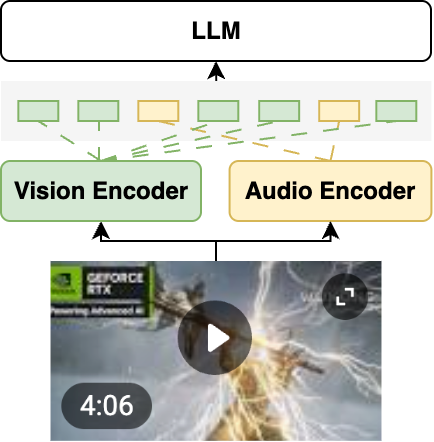}
    \centerline{(a)}
  \end{minipage}
  \hfill
  \begin{minipage}[t]{0.45\linewidth}
    \centering
    \includegraphics[width=\linewidth]{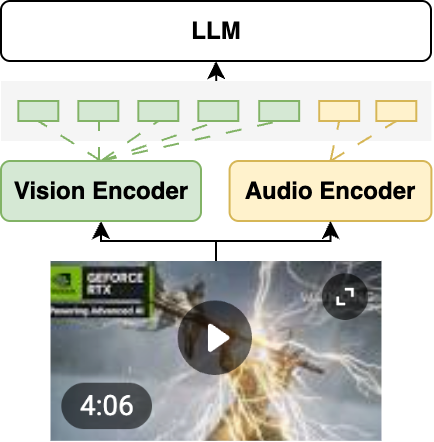}
    \centerline{(b)}
  \end{minipage}
  \caption{Comparison of two fusion strategies: (a) Qwen Omni model's interleaved fusion strategy, where audio and video tokens are organized sequentially and synchronized using TMRoPE. (b) Our retrieval model's separate-stream fusion strategy, where audio and video are encoded independently without token interleaving.}
  \Description{Two diagrams showing different fusion strategies for multimodal processing.}
  \label{fig:fusion}
\end{figure}

In contrast to the Qwen-Omni model, which interleaves audio and video tokens and applies Time-aligned Multimodal RoPE (TMRoPE) to enforce synchronous timestamps within a single token sequence~\cite{Qwen2.5-Omni} (Figure~\ref{fig:fusion} (a)), our retrieval encoder keeps the two streams separate (Figure~\ref{fig:fusion} (b)). We encode audio and video independently, preserving each modality’s native temporal structure without cross-modal token interleaving. According to our experiments, this non-interleaved design improves retrieval performance, as reported in Table~\ref{tab:multimodal_performance_lpm} and \ref{tab:multimodal_performance_finevideo}. We believe that for retrieval tasks, unlike generative tasks, keeping audio and video information complete without splitting them enables the model to capture the full context of each modality, thereby making it easier to match and retrieve relevant content.

\section{Training}

\subsection{Contrastive learning}

We leverage contrastive learning to maximize the embedding similarity between the query and positive corpus, while minimizing the similarity between the query and negative corpus.
We adopt the InfoNCE contrastive loss~\cite{infonce} to train the model to distinguish between positive and negative pairs in a shared embedding space:

\begin{equation}
\mathcal{L}(q, d^+, D_N) = -\log \frac{\exp(\text{sim}(q, d^+)/\tau)}{\sum_{d_i \in \{d^+\} \cup D_N} \exp(\text{sim}(q, d_i)/\tau)}
\label{eq:infonce}
\end{equation}   

\noindent where $q$ is the embedding of a query, and $d^+$ are embeddings positive documents. $D_N$ denotes the set of negative corpus. $ sim(\cdot) $ represents the similarity function (e.g., cosine similarity or dot product). $ \tau $ is the temperature parameter. 

To improve the effectiveness of contrastive learning, we incorporate the \textit{top-k with percentage to positive threshold} strategy from NV-Retriever~\cite{nv-retriever} for hard negative mining. We set the threshold as 0.95, meaning we select the $K$ most relevant negative samples whose similarity to the query is less than 95\% of the query–positive similarity score. We set $K=2$ in our experiments. This encourages the model to learn from challenging negatives, while removing potential false negatives that have high similarity scores.

\subsection{Training}
During training, we freeze both the audio and visual encoders, and apply LoRA~\cite{hu2022lora} tuning only to the language model (LLM). This approach allows us to efficiently adapt the model's cross-modal capabilities while minimizing computational overhead, effectively generalizing its ability to handle multimodal inputs. 
We set the hyperparameters LoRA rank and LoRA scaling factor as 
$r=16$ and  $\alpha=32$, respectively, to balance adaptation capacity with training efficiency.

We modify the model architecture by replacing causal attention with bidirectional attention and train it on large-scale text-text and text–image pairs. This helps establish strong foundational retrieval models for
queries and documents and align the learned representations with visual inputs, enabling alignment across different modalities.

\subsection{Datasets}
Following the methodology of Llama Nemoretriever Colembed~\cite{nemoretriever-colembed}, we fine-tune the model using a mixture of text–image pairs and high-quality, small-scale text-text retrieval datasets, including ColPali train set~\cite{colpali}, Wiki-SS-NQ~\cite{wikissnq, tevatron}, VDR~\cite{vdr},  Docmatix~\cite{docmatix}, Natural Questions (NQ)~\cite{nq}, Stack Exchange~\cite{stack}, SQuAD~\cite{squad} and HotpotQA~\cite{hotpotqa}. These diverse and challenging examples facilitate effective multimodal alignment, particularly for visual retrieval tasks. 

Even without using audio or video data during training, our model demonstrates strong performance on these modalities. To further extend its cross-modal capabilities, we experimented with an additional training stage that incorporated text–video and text–audio pairs from the FineVideo dataset~\cite{fineVideo}. Adding the FineVideo training set improved performance on the FineVideo evaluation benchmark, the experiments' results are presented in Table~\ref{tab:finevideo_train}. However, it did not enhance retrieval across other modalities or generalize better to unseen video datasets. Overall, the model’s unified embedding space for text, image, video, and audio remains largely unaffected by this additional stage.

\section{Results}

We evaluate our model across three modalities: video retrieval, image retrieval, and text-only retrieval. 
%This section first presents the results for the video retrieval task.

\subsection{Video Retrieval}

\paragraph{FineVideo Benchmark.}
FineVideo~\cite{fineVideo} contains 43,751 videos with an average duration of 4.7 minutes, spanning 122 diverse categories. To construct a realistic retrieval benchmark, we randomly sampled 10,000 videos as the retrieval corpus, while ensuring that the class distribution of content\_fine\_category is preserved in the test set. These videos include 46,553 associated questions, which we use as text queries for the retrieval task. Each query is paired with a ground-truth video from the corpus.

\paragraph{LPM Benchmark.}
We also constructed a second benchmark from the LPM dataset~\cite{lpm}, which focuses on educational content. The videos in LPM feature recorded lectures composed of live slide decks and the speaker’s head, with manually annotated slide segmentation timestamps. 
The dataset provides Google ASR transcriptions of the spoken content. We align these transcripts with their corresponding slide images and use these paired text–image inputs to prompt Llama-3.2-90B-Vision-Instruct~\cite{llama3.2-90b-vision}, generating 1,000 synthetic questions for evaluation.

\paragraph{Baselines.}

%to benchmark against strong text-only baselines, we use the speech-to-text transcripts provided in FineVideo and the transcripts we generated from LPM as the input corpus for standard text retrieval models. For comparison, we select Qwen/Qwen3-Embedding-4B~\cite{qwen3-embed}, the top-ranked model under 4B parameters on the MTEB leaderboard~\cite{mteb}, along with two other high-performing retrieval models: intfloat/multilingual-e5-large-instruct\cite{e5-multilingual} and stella\_en\_1.5B\_v5\cite{stella}. This evaluation setup enables a direct comparison between our multimodal retrieval model and state-of-the-art text-only retrieval systems operating solely on transcribed video content.

Because there is no standard text–video retrieval benchmark like ViDoRe for text-image retrieval or MTEB for text-text retrieval, we construct our own evaluation setup using speech-to-text transcripts provided in FineVideo and the transcripts we generated from LPM as the input corpus for standard text retrieval models. For the baselines, we select Qwen/Qwen3-Embedding-4B~\cite{qwen3-embed}, the top-ranked model under 4B parameters on the MTEB leaderboard~\cite{mteb}, along with two other high-performing retrieval models: intfloat/multilingual-e5-large-instruct~\cite{e5-multilingual} and stella\_en\_1.5B\_v5~\cite{stella}. This setup enables a direct comparison between our multimodal retrieval model and state-of-the-art text-only retrieval systems operating solely on transcribed video content.

\begin{table*}
  \caption{Model performance comparison on LPM and FineVideo datasets using NDCG@10 and NDCG@5 metrics. The Avg column reports the mean of LPM and FineVideo. Best averages are highlighted in bold. Rows 1–3 are text-only baselines evaluated on transcripts; row 4 is our omni model evaluating on video}
  \label{tab:video-results}
  \begin{tabular}{lcccccc}
    \toprule
    \multicolumn{1}{c}{{Model}} 
    & \multicolumn{3}{c}{NDCG@10} & \multicolumn{3}{c}{NDCG@5} \\
    \cmidrule(lr){2-4} \cmidrule(lr){5-7}
    & LPM & FineVideo & Avg & LPM & FineVideo & Avg \\
    \midrule
    Qwen/Qwen3-Embedding-4B 
      & 0.8634 & 0.5405 & 0.7020 & 0.8518 & 0.5264 & 0.6891 \\
    intfloat/multilingual-e5-large-instruct 
      & 0.7952 & 0.4456 & 0.6204 & 0.7759 & 0.4300 & 0.6030 \\
    stella\_en\_1.5B\_v5 
      & 0.8522 & 0.5359 & 0.6941 & 0.8404 & 0.5206 & 0.6805 \\
    nvidia/omni-embed-nemotron-3b 
      & 0.8465 & 0.5662 & \textbf{0.7064} & 0.8355 & 0.5486 & \textbf{0.6921} \\
    \bottomrule
  \end{tabular}
\end{table*}

\paragraph{Results.}
Table~\ref{tab:video-results} summarizes the retrieval performance of our model and baselines across both benchmarks. 
Across both datasets, our model is strongest on FineVideo, achieving the top NDCG@10 of 0.5662 and exceeding the best text-only baseline (Qwen3-Emb- edding-4B) by 2.6 points, respectively (Table~\ref{tab:video-results}). This indicates that leveraging multimodal cues (visual frames and audio) is beneficial when ASR transcripts are noisy or incomplete. On LPM’s lecture-style videos, where queries are synthesized from transcripts and text quality is high, Qwen3-Embedding-4B leads (0.8634/0.8518), while our model remains competitive at 0.8465/0.8355, trailing by 1.7 and 1.6 points and slightly behind stella\_en\_1.5B\_v5 but ahead of intfloat/multilingual-e5-large-instruct. Overall, multimodal retrieval yields clear gains in open-domain video, whereas in text-centric scenarios the advantage narrows. When looking at the averages, our model slightly outperforms the other models, reaching 0.7064 for NDCG@10 and 0.6921 for NDCG@5.

\paragraph{Modality Breakdown.}
Table~\ref{tab:multimodal_performance_lpm} and \ref{tab:multimodal_performance_finevideo} show performance on different input modalities. On LPM, using transcripts+OCR with our model (0.8636) essentially matches the best text baseline (0.8634), while audio-only (0.8238) and video-only (0.7365) trail; fusing audio+video improves to 0.8373, and scoring the two streams separately with late fusion reaches 0.8465, narrowing the gap to text. On FineVideo, our text-only variant (0.6082) surpasses all text baselines; among non-text channels, audio-only (0.5407) clearly outperforms video-only (0.4488). Simple early fusion of audio+video underperforms audio alone (0.4700), but late fusion (“Audio+Video Separately”) recovers to 0.5662, indicating the importance of the fusion strategy. Overall, text dominates in lecture-style content, whereas in open-domain videos audio carries more signal than frames alone, and late fusion yields the best non-text robustness when text is imperfect or unavailable.

% ---------- LPM TABLE ----------
\begin{table*}
  \caption{Multimodal retrieval performance on LPM dataset (NDCG@10). Current baseline models only support text input, while multimodal capabilities (Audio-Only, Video-Only, Audio+Video) are not available.}
  \label{tab:multimodal_performance_lpm}
  \begin{tabular}{lccccc}
    \toprule
    Model & \makecell{Text\\(Transcript+OCR)} & Audio-Only & Video-Only & \makecell{Audio+Video\\Fusion} & \makecell{Audio+Video\\Separately} \\
    \midrule
    Qwen/Qwen3-Embedding-4B & 0.8634 & N/A & N/A & N/A & N/A \\
    intfloat/multilingual-e5-large-instruct & 0.7952 & N/A & N/A & N/A & N/A \\
    stella\_en\_1.5B\_v5 & 0.8522 & N/A & N/A & N/A & N/A \\
    Omni-Embed-Nemotron-3B & 0.8636 & 0.8238 & 0.7365 & 0.8373 & 0.8465 \\
    \bottomrule
  \end{tabular}
\end{table*}

% ---------- FINEVIDEO TABLE ----------
\begin{table*}
  \caption{Multimodal retrieval performance on FineVideo dataset (NDCG@10). Current baseline models only support text input, while multimodal capabilities (Audio-Only, Video-Only, Audio+Video) are not available.}
  \label{tab:multimodal_performance_finevideo}
  \begin{tabular}{lccccc}
    \toprule
    Model & \makecell{Text\\(Transcript)} & Audio-Only & Video-Only & \makecell{Audio+Video\\Fusion} & \makecell{Audio+Video\\Separately} \\
    \midrule
    Qwen/Qwen3-Embedding-4B & 0.5405 & N/A & N/A & N/A & N/A \\
    intfloat/multilingual-e5-large-instruct & 0.4456 & N/A & N/A & N/A & N/A \\
    stella\_en\_1.5B\_v5 & 0.5359 & N/A & N/A & N/A & N/A \\
    Omni-Embed-Nemotron-3B & 0.6082 & 0.5407 & 0.4488 & 0.4700 & 0.5662 \\
    \bottomrule
  \end{tabular}
\end{table*}

\paragraph{Domain Specific Fine-Tuning}

Although the model is not explicitly trained on any video datasets, the results demonstrate strong performance on the video retrieval task, highlighting the model’s generalization capability across modalities.

To further assess the benefits of domain-specific adaptation, we fine-tune the model on the training set of FineVideo and evaluate it on the corresponding test set. The dataset is randomly split at the video level, ensuring that no videos overlap between training and evaluation. This fine-tuning leads to a performance gain as shown in Table~\ref{tab:finevideo_train}, with ndcg@5 improving from 0.55 to 0.61. This result confirms that task-specific fine-tuning on in-domain data can effectively enhance retrieval accuracy in the video domain.

\begin{table}[t]
  \caption{Video retrieval performance on FineVideo test set. Our model demonstrates strong zero-shot generalization to video domain, while domain-specific fine-tuning provides additional performance gains.}
  \label{tab:finevideo_train}
  \centering
  \begin{tabular}{lcc}
    \toprule
    Model Configuration & NDCG@10 & NDCG@5 \\
    \midrule
    \multicolumn{3}{l}{\textit{Baseline Models}} \\
    Qwen/Qwen3-Embedding-4B & 0.5405 & 0.5264 \\
    intfloat/multilingual-e5-large-instruct & 0.4456 & 0.4300 \\
    stella\_en\_1.5B\_v5 & 0.5359 & 0.5206 \\
    \midrule
    \multicolumn{3}{l}{\textit{Our Models}} \\
    Omni-Embed-Nemotron-3B & 0.5662 & 0.5486 \\
    + Fine-tuned on FineVideo training set & 0.6251 & 0.6092 \\
    \bottomrule
  \end{tabular}
\end{table}

%{'NDCG@1': 0.52486, 'NDCG@5': 0.60923, 'NDCG@10': 0.62506, 'NDCG@100': 0.65224} {'MAP@1': 0.52478, 'MAP@5': 0.5851, 'MAP@10': 0.59164, 'MAP@100': 0.5969} {'Recall@1': 0.52478, 'Recall@5': 0.68124, 'Recall@10': 0.73011, 'Recall@100': 0.86055} {'P@1': 0.52486, 'P@5': 0.13635, 'P@10': 0.07309, 'P@100': 0.00865}

%we further fine-tune the model on cross-modal pairs involving text-video, text-audio, and text-image data. This stage enables the model to handle complex cross-modal retrieval scenarios by learning fine-grained alignment across diverse modalities. It also ensures that the shared embedding space supports flexible query-document combinations, such as retrieving a video using an audio-text query or an image using a text-only query.

\subsection{Image Retrieval}

We evaluate our model’s image retrieval performance using the ViDoRe benchmark~\cite{colpali}, a large-scale multimodal benchmark designed for evaluating visual understanding and retrieval. ViDoRe consists of a diverse collection of images paired with natural language queries, making it suitable for testing cross-modal alignment between text and image representations.

To establish strong baselines, we compare our model with several high-performing image retrieval models under the 4B parameter scale. Notably, we include nvidia/llama-nemoretriever-colembed-1b-v1~\cite{nemoretriever-colembed}, the top-ranked model among all models under 4B parameters in this category. This model represents the current state of the art in lightweight multimodal retrieval.

Results of this evaluation are presented in Table~\ref{tab:image-results}, the leading baseline (\textit{llama-nemoretriever-colembed-3B}) averages 91.0, while our omni model averages 85.7. Our model is competitive on ArxivQA and strong on broad domains (AI, Government Reports, Energy), but underperform on Shift Project, TAT-DQA, DocVQA. Overall, the results indicate that our model has strong capability to retrieve visual documents.

\begin{table*}[t]
  \caption{Evaluation of baseline models and our models on 
  \href{https://huggingface.co/spaces/vidore/vidore-leaderboard}{ViDoRe V1} (as of July 30th). 
  Results are presented using nDCG@5 metrics.}
  \label{tab:image-results}
  \centering
  \resizebox{\textwidth}{!}{%
    \begin{tabular}{lcccccccccccc}
      \toprule
      Model & Size (M) & Avg. & ArxivQA & DocVQA & InfoVQA & Shift Project & AI & Energy & Gov. Reports & Healthcare & TabFQuad & TAT-DQA \\
      \midrule
      nvidia/llama-nemoretriever-colembed-1b-v1
        & 2418 & 90.5 & 87.6 & 64.5 & 93.6 & 92.3 & 100 & 96.6 & 96.7 & 99.6 & 94.3 & 79.8 \\
      nvidia/llama-nemoretriever-colembed-3b-v1
        & 4407 & 91.0 & 88.4 & 66.2 & 94.9 & 90.7 & 99.6 & 96.6 & 97.8 & 99.3 & 95.9 & 80.6 \\
      nomic-ai/colnomic-embed-multimodal-3b
        & 3000 & 89.9 & 88.2 & 61.3 & 92.8 & 90.2 & 96.3 & 97.3 & 96.6 & 98.3 & 94.5 & 83.1 \\
      vidore/colqwen2.5-v0.2
        & 3000 & 89.6 & 89.1 & 63.5 & 92.6 & 88.0 & 99.6 & 95.8 & 96.6 & 98.0 & 90.8 & 82.1 \\
      vidore/colqwen2-v1.0
        & 2210 & 89.2 & 88.0 & 61.5 & 92.5 & 89.9 & 99.0 & 95.9 & 95.5 & 98.8 & 89.0 & 82.2 \\
      vidore/colpali-v1.3
        & 2920 & 84.7 & 83.7 & 58.7 & 85.7 & 76.5 & 96.6 & 94.6 & 95.9 & 97.4 & 86.7 & 70.7 \\
      vidore/colpali-v1.2
        & 2920 & 83.4 & 77.9 & 56.5 & 82.4 & 78.3 & 97.5 & 94.4 & 94.9 & 95.4 & 88.4 & 68.1 \\
      \midrule
      \textbf{Ours} & & & & & & & & & & & \\
      nvidia/omni-embed-nemotron-3b
        & 4703 & 85.7 & 85.3 & 59.2 & 89.2 & 78.6 & 98.1 & 93.5 & 95.4 & 95.8 & 91.0 & 69.7 \\
      \bottomrule
    \end{tabular}
  }
\end{table*}

\subsection{Text Retrieval}
To evaluate the text retrieval capability of our model, we select several benchmark tasks from the Massive Text Embedding Benchmark (MTEB)~\cite{mteb}. These tasks cover a variety of domains such as open-domain question answering, scientific literature retrieval, and argumentative text matching, providing a broad view of how the model performs across different retrieval scenarios.

As shown in Table~\ref{tab:text-retrieval}, although the model does not achieve the strongest results compared to specialized text-only embeddings, it delivers solid and competitive performance across datasets.
Most importantly, the results show that the model can serve as a general-purpose retrieval system across modalities, maintaining good performance on text while also supporting image, video, and audio retrieval within a unified embedding space.

\begin{table*}[t]
  \caption{Evaluation of embedding models across text retrieval benchmarks. Results are reported using nDCG@10.}
  \label{tab:text-retrieval}
  \centering
  \resizebox{\textwidth}{!}{%
    \begin{tabular}{lcccccccccccc}
      \toprule
      Model & Avg. & NQ & FiQA-2018 & SciFact & SCIDOCS & ArguAna & NFCorpus & Quora & LegalBench-CorpLobby & CQAdupGaming & CQAdupUnix \\
      \midrule
      Qwen/Qwen3-Embedding-4B
        & 0.6654 & 0.6313 & 0.6122 & 0.7833 & 0.3144 & 0.7564 & 0.4110 & 0.8806 & 0.9542 & 0.7151 & 0.5960 \\
      intfloat/multilingual-e5-large-instruct
        & 0.5900 & 0.6350 & 0.4865 & 0.7162 & 0.1924 & 0.5848 & 0.3634 & 0.8926 & 0.9425 & 0.6396 & 0.4473 \\
      stella\_en\_1.5B\_v5
        & 0.6050 & 0.7180 & 0.5996 & 0.8009 & 0.2677 & 0.5706 & 0.4200 & 0.9003 & 0.9468 & 0.5359 & 0.2903 \\
      nvidia/omni-embed-nemotron-3b
        & 0.6059 & 0.6808 & 0.5382 & 0.7405 & 0.2163 & 0.5891 & 0.3644 & 0.8347 & 0.9413 & 0.6432 & 0.5102 \\
      \bottomrule
    \end{tabular}
  }
\end{table*}

\section{Impact of Modality and Preprocessing on Sequence Length and Retrieval Performance}

%Our ablation study shows that the processor arguments strongly influence both the sequence length and the resulting retrieval performance. Careful configuration is necessary to balance computational cost and accuracy. Based on our experience, the configuration setting shown in Table~\ref{tab:processor_args} provides stable performance across different modalities. To illustrate, we take an MP4 video with a duration of 1050.67 seconds and a file size of 28 MB as an example. Using the above processor settings, we calculate its sequence length after preprocessing, which highlights how processor design choices directly determine the tokenized input length and thus the compute requirements of multimodal retrieval.

Our ablation study shows that the processor arguments strongly influence both the sequence length and the resulting retrieval performance. Careful configuration is necessary to balance computational cost and accuracy. Based on our experience, the configuration setting shown in Table~\ref{tab:processor_args} provides stable performance across different modalities. To illustrate, we take an MP4 video with a duration of 1050.67 seconds and a file size of 28 MB as an example. Using the processor settings in Table~\ref{tab:processor_args}, we calculate its sequence length after preprocessing, which highlights how processor design choices directly determine the tokenized input length and thus the compute requirements of multimodal retrieval. Furthermore, modality selection impacts retrieval performance: combining video and audio yields better results than using either modality alone, as shown in Table~\ref{tab:multimodal_performance_lpm} and \ref{tab:multimodal_performance_finevideo}. Alternatively, applying ASR to extract text from videos can be efficient for content-rich audio like Finevideo, but this comes at the cost of additional ASR processing. Overall, these findings emphasize the importance of modality-aware preprocessing for building efficient and scalable retrieval systems.

% ---------- Table 1 ----------
\begin{table}[t]
  \caption{Processor argument settings used in our ablation study. These values balance sequence length and performance across different modalities.}
  \label{tab:processor_args}
  \centering
  \begin{tabular}{lc}
    \toprule
    \textbf{Processor Argument} & \textbf{Value} \\
    \midrule
    \texttt{min\_pixels} & $32 \times 14 \times 14$ \\
    \texttt{max\_pixels} & $64 \times 28 \times 28$ \\
    \texttt{audio\_kwargs.max\_length} & $2{,}048{,}000$ \\
    \texttt{image\_max\_pixels} & $2352$ \\
    \texttt{image\_min\_pixels} & $196$ \\
    \bottomrule
  \end{tabular}
\end{table}

% ---------- Table 2 ----------
\begin{table}[t]
  \caption{Sequence lengths under different modality settings. The results indicate that different modalities lead to varying sequence lengths, which directly impact compute requirements during retrieval.}
  \label{tab:seq_len_ablation}
  \centering
  \begin{tabular}{lc}
    \toprule
    \textbf{Modality Setting} & \textbf{Sequence Length} \\
    \midrule
    Text (Transcript + OCR) & $3497$ \\
    Audio-Only & $3222$ \\
    Video-Only & $20758$ \\
    Audio + Video (Fusion) & $23960$ \\
    Audio + Video (Separately) & $23960$ \\
    \bottomrule
  \end{tabular}
\end{table}

%adding pretrain

\section{Conclusion}
We presented Omni-Embed-Nemotron-3B, a multimodal retrieval model built on the Qwen-Omni backbone that is capable of handling text, image, audio, and video in a unified embedding space. Through architectural simplifications, modality-specific encoding, and contrastive training, our model achieves strong performance across multiple benchmarks. While not always outperforming specialized single-modality systems, it demonstrates robust and competitive results in text, image, and video retrieval. These findings highlight the effectiveness of a unified retrieval model for diverse modalities, and point toward future improvements through domain-specific fine-tuning and more sophisticated multimodal alignment strategies.

\bibliographystyle{ACM-Reference-Format}
\bibliography{omni-embed-sigconf}


\begin{thebibliography}{32}

%%% the Web address.

\ifx \showCODEN    \undefined \def \showCODEN     #1{\unskip}     \fi
\ifx \showISBNx    \undefined \def \showISBNx     #1{\unskip}     \fi
\ifx \showISBNxiii \undefined \def \showISBNxiii  #1{\unskip}     \fi
\ifx \showISSN     \undefined \def \showISSN      #1{\unskip}     \fi
\ifx \showLCCN     \undefined \def \showLCCN      #1{\unskip}     \fi
\ifx \shownote     \undefined \def \shownote      #1{#1}          \fi
\ifx \showarticletitle \undefined \def \showarticletitle #1{#1}   \fi
\ifx \showURL      \undefined \def \showURL       {\relax}        \fi
% invisible to TeX
\providecommand\bibfield[2]{#2}
\providecommand\bibinfo[2]{#2}
\providecommand\natexlab[1]{#1}
\providecommand\showeprint[2][]{arXiv:#2}

\bibitem[AI(2024)]%
        {llama3.2-90b-vision}
\bibfield{author}{\bibinfo{person}{Meta AI}.} \bibinfo{year}{2024}\natexlab{}.
\newblock \bibinfo{title}{LLaMA 3.2 90B Vision Instruct}.
\newblock \bibinfo{howpublished}{\url{https://huggingface.co/meta-llama/Llama-3.2-90B-Vision}}.
\newblock
\newblock
\shownote{Instruction-tuned multimodal model with vision and language capabilities, released by Meta AI in September 2024}.


\bibitem[Bercovich et~al\mbox{.}(2025)]%
        {nemotron}
\bibfield{author}{\bibinfo{person}{Akhiad Bercovich}, \bibinfo{person}{Itay Levy}, \bibinfo{person}{Izik Golan}, \bibinfo{person}{Mohammad Dabbah}, \bibinfo{person}{Ran El-Yaniv}, \bibinfo{person}{Omri Puny}, \bibinfo{person}{Ido Galil}, \bibinfo{person}{Zach Moshe}, \bibinfo{person}{Tomer Ronen}, \bibinfo{person}{Najeeb Nabwani}, {et~al\mbox{.}}} \bibinfo{year}{2025}\natexlab{}.
\newblock \showarticletitle{Llama-nemotron: Efficient reasoning models}.
\newblock \bibinfo{journal}{\emph{arXiv preprint arXiv:2505.00949}} (\bibinfo{year}{2025}).
\newblock


\bibitem[Chen et~al\mbox{.}(2025)]%
        {eagle25}
\bibfield{author}{\bibinfo{person}{Guo Chen}, \bibinfo{person}{Zhiqi Li}, \bibinfo{person}{Shihao Wang}, \bibinfo{person}{Jindong Jiang}, \bibinfo{person}{Yicheng Liu}, \bibinfo{person}{Lidong Lu}, \bibinfo{person}{De-An Huang}, \bibinfo{person}{Wonmin Byeon}, \bibinfo{person}{Matthieu Le}, \bibinfo{person}{Tuomas Rintamaki}, {et~al\mbox{.}}} \bibinfo{year}{2025}\natexlab{}.
\newblock \showarticletitle{Eagle 2.5: Boosting long-context post-training for frontier vision-language models}.
\newblock \bibinfo{journal}{\emph{arXiv preprint arXiv:2504.15271}} (\bibinfo{year}{2025}).
\newblock


\bibitem[Chen et~al\mbox{.}(2020)]%
        {infonce}
\bibfield{author}{\bibinfo{person}{Ting Chen}, \bibinfo{person}{Simon Kornblith}, \bibinfo{person}{Mohammad Norouzi}, {and} \bibinfo{person}{Geoffrey Hinton}.} \bibinfo{year}{2020}\natexlab{}.
\newblock \showarticletitle{A simple framework for contrastive learning of visual representations}. In \bibinfo{booktitle}{\emph{International conference on machine learning}}. PmLR, \bibinfo{pages}{1597--1607}.
\newblock


\bibitem[Chung et~al\mbox{.}(2025)]%
        {mtebnew}
\bibfield{author}{\bibinfo{person}{Isaac Chung}, \bibinfo{person}{Imene Kerboua}, \bibinfo{person}{Marton Kardos}, \bibinfo{person}{Roman Solomatin}, {and} \bibinfo{person}{Kenneth Enevoldsen}.} \bibinfo{year}{2025}\natexlab{}.
\newblock \showarticletitle{Maintaining MTEB: Towards Long Term Usability and Reproducibility of Embedding Benchmarks}.
\newblock \bibinfo{journal}{\emph{arXiv preprint arXiv:2506.21182}} (\bibinfo{year}{2025}).
\newblock


\bibitem[Cimolai and Markewich(2025)]%
        {vdr}
\bibfield{author}{\bibinfo{person}{Marco Cimolai} {and} \bibinfo{person}{Logan Markewich}.} \bibinfo{year}{2025}\natexlab{}.
\newblock \bibinfo{title}{llamaindex/vdr-multilingual-train}.
\newblock
\urldef\tempurl%
\url{https://huggingface.co/datasets/llamaindex/vdr-multilingual-train}
\showURL{%
Retrieved June 30, 2025 from \tempurl}


\bibitem[Farré et~al\mbox{.}(2024)]%
        {fineVideo}
\bibfield{author}{\bibinfo{person}{Miquel Farré}, \bibinfo{person}{Andi Marafioti}, \bibinfo{person}{Lewis Tunstall}, \bibinfo{person}{Leandro Von~Werra}, {and} \bibinfo{person}{Thomas Wolf}.} \bibinfo{year}{2024}\natexlab{}.
\newblock \bibinfo{title}{FineVideo}.
\newblock \bibinfo{howpublished}{\url{https://huggingface.co/datasets/HuggingFaceFV/finevideo}}.
\newblock


\bibitem[Faysse et~al\mbox{.}(2024)]%
        {colpali}
\bibfield{author}{\bibinfo{person}{Manuel Faysse}, \bibinfo{person}{Hugues Sibille}, \bibinfo{person}{Tony Wu}, \bibinfo{person}{Bilel Omrani}, \bibinfo{person}{Gautier Viaud}, \bibinfo{person}{Céline Hudelot}, {and} \bibinfo{person}{Pierre Colombo}.} \bibinfo{year}{2024}\natexlab{}.
\newblock \bibinfo{title}{ColPali: Efficient Document Retrieval with Vision Language Models}.
\newblock
\showeprint[arxiv]{2407.01449}~[cs.IR]
\urldef\tempurl%
\url{https://arxiv.org/abs/2407.01449}
\showURL{%
\tempurl}


\bibitem[Hu et~al\mbox{.}(2022)]%
        {hu2022lora}
\bibfield{author}{\bibinfo{person}{Edward~J Hu}, \bibinfo{person}{Yelong Shen}, \bibinfo{person}{Phillip Wallis}, \bibinfo{person}{Zeyuan Allen-Zhu}, \bibinfo{person}{Yuanzhi Li}, \bibinfo{person}{Shean Wang}, \bibinfo{person}{Lu Wang}, \bibinfo{person}{Weizhu Chen}, {et~al\mbox{.}}} \bibinfo{year}{2022}\natexlab{}.
\newblock \showarticletitle{Lora: Low-rank adaptation of large language models.}
\newblock \bibinfo{journal}{\emph{ICLR}} \bibinfo{volume}{1}, \bibinfo{number}{2} (\bibinfo{year}{2022}), \bibinfo{pages}{3}.
\newblock


\bibitem[Jin~Xu(2025)]%
        {Qwen2.5-Omni}
\bibfield{author}{\bibinfo{person}{Jinzheng~He Jin~Xu, Zhifang~Guo}.} \bibinfo{year}{2025}\natexlab{}.
\newblock \showarticletitle{Qwen2.5-Omni Technical Report}.
\newblock \bibinfo{journal}{\emph{arXiv preprint arXiv:2503.20215}} (\bibinfo{year}{2025}).
\newblock


\bibitem[Kwiatkowski et~al\mbox{.}(2019)]%
        {nq}
\bibfield{author}{\bibinfo{person}{Tom Kwiatkowski}, \bibinfo{person}{Jennimaria Palomaki}, \bibinfo{person}{Olivia Redfield}, \bibinfo{person}{Michael Collins}, \bibinfo{person}{Ankur Parikh}, \bibinfo{person}{Chris Alberti}, \bibinfo{person}{Danielle Epstein}, \bibinfo{person}{Illia Polosukhin}, \bibinfo{person}{Jacob Devlin}, \bibinfo{person}{Kenton Lee}, {et~al\mbox{.}}} \bibinfo{year}{2019}\natexlab{}.
\newblock \showarticletitle{Natural questions: a benchmark for question answering research}.
\newblock \bibinfo{journal}{\emph{Transactions of the Association for Computational Linguistics}}  \bibinfo{volume}{7} (\bibinfo{year}{2019}), \bibinfo{pages}{453--466}.
\newblock


\bibitem[Laurençon et~al\mbox{.}(2024)]%
        {docmatix}
\bibfield{author}{\bibinfo{person}{Hugo Laurençon}, \bibinfo{person}{Andrés Marafioti}, \bibinfo{person}{Victor Sanh}, {and} \bibinfo{person}{Léo Tronchon}.} \bibinfo{year}{2024}\natexlab{}.
\newblock \bibinfo{title}{Building and better understanding vision-language models: insights and future directions.}
\newblock
\showeprint[arxiv]{2408.12637}~[cs.CV]


\bibitem[Lee et~al\mbox{.}(2024)]%
        {nv-embed}
\bibfield{author}{\bibinfo{person}{Chankyu Lee}, \bibinfo{person}{Rajarshi Roy}, \bibinfo{person}{Mengyao Xu}, \bibinfo{person}{Jonathan Raiman}, \bibinfo{person}{Mohammad Shoeybi}, \bibinfo{person}{Bryan Catanzaro}, {and} \bibinfo{person}{Wei Ping}.} \bibinfo{year}{2024}\natexlab{}.
\newblock \showarticletitle{Nv-embed: Improved techniques for training llms as generalist embedding models}.
\newblock \bibinfo{journal}{\emph{arXiv preprint arXiv:2405.17428}} (\bibinfo{year}{2024}).
\newblock


\bibitem[Lee et~al\mbox{.}(2022)]%
        {lpm}
\bibfield{author}{\bibinfo{person}{Dong~Won Lee}, \bibinfo{person}{Chaitanya Ahuja}, \bibinfo{person}{Paul~Pu Liang}, \bibinfo{person}{Sanika Natu}, {and} \bibinfo{person}{Louis-Philippe Morency}.} \bibinfo{year}{2022}\natexlab{}.
\newblock \showarticletitle{Multimodal lecture presentations dataset: Understanding multimodality in educational slides}.
\newblock \bibinfo{journal}{\emph{arXiv preprint arXiv:2208.08080}} (\bibinfo{year}{2022}).
\newblock


\bibitem[Li et~al\mbox{.}(2025)]%
        {eagle}
\bibfield{author}{\bibinfo{person}{Zhiqi Li}, \bibinfo{person}{Guo Chen}, \bibinfo{person}{Shilong Liu}, \bibinfo{person}{Shihao Wang}, \bibinfo{person}{Vibashan VS}, \bibinfo{person}{Yishen Ji}, \bibinfo{person}{Shiyi Lan}, \bibinfo{person}{Hao Zhang}, \bibinfo{person}{Yilin Zhao}, \bibinfo{person}{Subhashree Radhakrishnan}, {et~al\mbox{.}}} \bibinfo{year}{2025}\natexlab{}.
\newblock \showarticletitle{Eagle 2: Building Post-Training Data Strategies from Scratch for Frontier Vision-Language Models}.
\newblock \bibinfo{journal}{\emph{arXiv preprint arXiv:2501.14818}} (\bibinfo{year}{2025}).
\newblock


\bibitem[Ma et~al\mbox{.}(2025)]%
        {tevatron}
\bibfield{author}{\bibinfo{person}{Xueguang Ma}, \bibinfo{person}{Luyu Gao}, \bibinfo{person}{Shengyao Zhuang}, \bibinfo{person}{Jiaqi~Samantha Zhan}, \bibinfo{person}{Jamie Callan}, {and} \bibinfo{person}{Jimmy Lin}.} \bibinfo{year}{2025}\natexlab{}.
\newblock \showarticletitle{Tevatron 2.0: Unified Document Retrieval Toolkit across Scale, Language, and Modality}.
\newblock \bibinfo{journal}{\emph{arXiv preprint arXiv:2505.02466}} (\bibinfo{year}{2025}).
\newblock


\bibitem[Moreira et~al\mbox{.}(2024)]%
        {nv-retriever}
\bibfield{author}{\bibinfo{person}{Gabriel de Souza~P Moreira}, \bibinfo{person}{Radek Osmulski}, \bibinfo{person}{Mengyao Xu}, \bibinfo{person}{Ronay Ak}, \bibinfo{person}{Benedikt Schifferer}, {and} \bibinfo{person}{Even Oldridge}.} \bibinfo{year}{2024}\natexlab{}.
\newblock \showarticletitle{NV-Retriever: Improving text embedding models with effective hard-negative mining}.
\newblock \bibinfo{journal}{\emph{arXiv preprint arXiv:2407.15831}} (\bibinfo{year}{2024}).
\newblock


\bibitem[Muennighoff et~al\mbox{.}(2022)]%
        {mteb}
\bibfield{author}{\bibinfo{person}{Niklas Muennighoff}, \bibinfo{person}{Nouamane Tazi}, \bibinfo{person}{Lo{\"\i}c Magne}, {and} \bibinfo{person}{Nils Reimers}.} \bibinfo{year}{2022}\natexlab{}.
\newblock \showarticletitle{MTEB: Massive text embedding benchmark}.
\newblock \bibinfo{journal}{\emph{arXiv preprint arXiv:2210.07316}} (\bibinfo{year}{2022}).
\newblock


\bibitem[Radford et~al\mbox{.}(2021)]%
        {clip}
\bibfield{author}{\bibinfo{person}{Alec Radford}, \bibinfo{person}{Jong~Wook Kim}, \bibinfo{person}{Chris Hallacy}, \bibinfo{person}{Aditya Ramesh}, \bibinfo{person}{Gabriel Goh}, \bibinfo{person}{Sandhini Agarwal}, \bibinfo{person}{Girish Sastry}, \bibinfo{person}{Amanda Askell}, \bibinfo{person}{Pamela Mishkin}, \bibinfo{person}{Jack Clark}, {et~al\mbox{.}}} \bibinfo{year}{2021}\natexlab{}.
\newblock \showarticletitle{Learning transferable visual models from natural language supervision}. In \bibinfo{booktitle}{\emph{International conference on machine learning}}. PmLR, \bibinfo{pages}{8748--8763}.
\newblock


\bibitem[Rajpurkar et~al\mbox{.}(2016)]%
        {squad}
\bibfield{author}{\bibinfo{person}{Pranav Rajpurkar}, \bibinfo{person}{Jian Zhang}, \bibinfo{person}{Konstantin Lopyrev}, {and} \bibinfo{person}{Percy Liang}.} \bibinfo{year}{2016}\natexlab{}.
\newblock \showarticletitle{Squad: 100,000+ questions for machine comprehension of text}.
\newblock \bibinfo{journal}{\emph{arXiv preprint arXiv:1606.05250}} (\bibinfo{year}{2016}).
\newblock


\bibitem[Ranzinger et~al\mbox{.}(2024)]%
        {cradio}
\bibfield{author}{\bibinfo{person}{Mike Ranzinger}, \bibinfo{person}{Greg Heinrich}, \bibinfo{person}{Jan Kautz}, {and} \bibinfo{person}{Pavlo Molchanov}.} \bibinfo{year}{2024}\natexlab{}.
\newblock \showarticletitle{Am-radio: Agglomerative vision foundation model reduce all domains into one}. In \bibinfo{booktitle}{\emph{Proceedings of the IEEE/CVF Conference on Computer Vision and Pattern Recognition}}. \bibinfo{pages}{12490--12500}.
\newblock


\bibitem[{Stack Exchange, Inc.}(2023)]%
        {stack}
\bibfield{author}{\bibinfo{person}{{Stack Exchange, Inc.}}} \bibinfo{year}{2023}\natexlab{}.
\newblock \bibinfo{title}{{Stack Exchange Community Data Dump, 2023}}.
\newblock \bibinfo{howpublished}{\url{https://archive.org/details/stack-exchange-data-dump-2023-09-12}}.
\newblock
\newblock
\shownote{Accessed: 2025-06-30}.


\bibitem[Team(2025)]%
        {nomic}
\bibfield{author}{\bibinfo{person}{Nomic Team}.} \bibinfo{year}{2025}\natexlab{}.
\newblock \bibinfo{title}{Nomic Embed Multimodal: Interleaved Text, Image, and Screenshots for Visual Document Retrieval}.
\newblock
\urldef\tempurl%
\url{https://nomic.ai/blog/posts/nomic-embed-multimodal}
\showURL{%
\tempurl}


\bibitem[Tschannen et~al\mbox{.}(2025)]%
        {siglip}
\bibfield{author}{\bibinfo{person}{Michael Tschannen}, \bibinfo{person}{Alexey Gritsenko}, \bibinfo{person}{Xiao Wang}, \bibinfo{person}{Muhammad~Ferjad Naeem}, \bibinfo{person}{Ibrahim Alabdulmohsin}, \bibinfo{person}{Nikhil Parthasarathy}, \bibinfo{person}{Talfan Evans}, \bibinfo{person}{Lucas Beyer}, \bibinfo{person}{Ye Xia}, \bibinfo{person}{Basil Mustafa}, {et~al\mbox{.}}} \bibinfo{year}{2025}\natexlab{}.
\newblock \showarticletitle{Siglip 2: Multilingual vision-language encoders with improved semantic understanding, localization, and dense features}.
\newblock \bibinfo{journal}{\emph{arXiv preprint arXiv:2502.14786}} (\bibinfo{year}{2025}).
\newblock


\bibitem[Wang et~al\mbox{.}(2023)]%
        {e5-mistral}
\bibfield{author}{\bibinfo{person}{Liang Wang}, \bibinfo{person}{Nan Yang}, \bibinfo{person}{Xiaolong Huang}, \bibinfo{person}{Linjun Yang}, \bibinfo{person}{Rangan Majumder}, {and} \bibinfo{person}{Furu Wei}.} \bibinfo{year}{2023}\natexlab{}.
\newblock \showarticletitle{Improving Text Embeddings with Large Language Models}.
\newblock \bibinfo{journal}{\emph{arXiv preprint arXiv:2401.00368}} (\bibinfo{year}{2023}).
\newblock


\bibitem[Wang et~al\mbox{.}(2024b)]%
        {e5-multilingual}
\bibfield{author}{\bibinfo{person}{Liang Wang}, \bibinfo{person}{Nan Yang}, \bibinfo{person}{Xiaolong Huang}, \bibinfo{person}{Linjun Yang}, \bibinfo{person}{Rangan Majumder}, {and} \bibinfo{person}{Furu Wei}.} \bibinfo{year}{2024}\natexlab{b}.
\newblock \showarticletitle{Multilingual e5 text embeddings: A technical report}.
\newblock \bibinfo{journal}{\emph{arXiv preprint arXiv:2402.05672}} (\bibinfo{year}{2024}).
\newblock


\bibitem[Wang et~al\mbox{.}(2024a)]%
        {qwenvl}
\bibfield{author}{\bibinfo{person}{Peng Wang}, \bibinfo{person}{Shuai Bai}, \bibinfo{person}{Sinan Tan}, \bibinfo{person}{Shijie Wang}, \bibinfo{person}{Zhihao Fan}, \bibinfo{person}{Jinze Bai}, \bibinfo{person}{Keqin Chen}, \bibinfo{person}{Xuejing Liu}, \bibinfo{person}{Jialin Wang}, \bibinfo{person}{Wenbin Ge}, {et~al\mbox{.}}} \bibinfo{year}{2024}\natexlab{a}.
\newblock \showarticletitle{Qwen2-vl: Enhancing vision-language model's perception of the world at any resolution}.
\newblock \bibinfo{journal}{\emph{arXiv preprint arXiv:2409.12191}} (\bibinfo{year}{2024}).
\newblock


\bibitem[Xu et~al\mbox{.}(2025)]%
        {nemoretriever-colembed}
\bibfield{author}{\bibinfo{person}{Mengyao Xu}, \bibinfo{person}{Gabriel Moreira}, \bibinfo{person}{Ronay Ak}, \bibinfo{person}{Radek Osmulski}, \bibinfo{person}{Yauhen Babakhin}, \bibinfo{person}{Zhiding Yu}, \bibinfo{person}{Benedikt Schifferer}, {and} \bibinfo{person}{Even Oldridge}.} \bibinfo{year}{2025}\natexlab{}.
\newblock \showarticletitle{Llama Nemoretriever Colembed: Top-Performing Text-Image Retrieval Model}.
\newblock \bibinfo{journal}{\emph{arXiv preprint arXiv:2507.05513}} (\bibinfo{year}{2025}).
\newblock


\bibitem[Xueguang~Ma(2024)]%
        {wikissnq}
\bibfield{author}{\bibinfo{person}{et~al. Xueguang~Ma}.} \bibinfo{year}{2024}\natexlab{}.
\newblock \bibinfo{title}{Tevatron/wiki-ss-nq}.
\newblock
\urldef\tempurl%
\url{https://huggingface.co/datasets/Tevatron/wiki-ss-nq}
\showURL{%
Retrieved June 30, 2025 from \tempurl}


\bibitem[Yang et~al\mbox{.}(2018)]%
        {hotpotqa}
\bibfield{author}{\bibinfo{person}{Zhilin Yang}, \bibinfo{person}{Peng Qi}, \bibinfo{person}{Saizheng Zhang}, \bibinfo{person}{Yoshua Bengio}, \bibinfo{person}{William~W Cohen}, \bibinfo{person}{Ruslan Salakhutdinov}, {and} \bibinfo{person}{Christopher~D Manning}.} \bibinfo{year}{2018}\natexlab{}.
\newblock \showarticletitle{HotpotQA: A dataset for diverse, explainable multi-hop question answering}.
\newblock \bibinfo{journal}{\emph{arXiv preprint arXiv:1809.09600}} (\bibinfo{year}{2018}).
\newblock


\bibitem[Zhang et~al\mbox{.}(2025b)]%
        {stella}
\bibfield{author}{\bibinfo{person}{Dun Zhang}, \bibinfo{person}{Jiacheng Li}, \bibinfo{person}{Ziyang Zeng}, {and} \bibinfo{person}{Fulong Wang}.} \bibinfo{year}{2025}\natexlab{b}.
\newblock \bibinfo{title}{Jasper and Stella: distillation of SOTA embedding models}.
\newblock
\showeprint[arxiv]{2412.19048}~[cs.IR]
\urldef\tempurl%
\url{https://arxiv.org/abs/2412.19048}
\showURL{%
\tempurl}


\bibitem[Zhang et~al\mbox{.}(2025a)]%
        {qwen3-embed}
\bibfield{author}{\bibinfo{person}{Yanzhao Zhang}, \bibinfo{person}{Mingxin Li}, \bibinfo{person}{Dingkun Long}, \bibinfo{person}{Xin Zhang}, \bibinfo{person}{Huan Lin}, \bibinfo{person}{Baosong Yang}, \bibinfo{person}{Pengjun Xie}, \bibinfo{person}{An Yang}, \bibinfo{person}{Dayiheng Liu}, \bibinfo{person}{Junyang Lin}, \bibinfo{person}{Fei Huang}, {and} \bibinfo{person}{Jingren Zhou}.} \bibinfo{year}{2025}\natexlab{a}.
\newblock \showarticletitle{Qwen3 Embedding: Advancing Text Embedding and Reranking Through Foundation Models}.
\newblock \bibinfo{journal}{\emph{arXiv preprint arXiv:2506.05176}} (\bibinfo{year}{2025}).
\newblock


\end{thebibliography}

\end{document}